\documentclass[conference]{IEEEtran}
\usepackage{amsmath,amssymb,amsfonts}
\usepackage{algorithmic}
\usepackage{graphicx}
\usepackage{textcomp}
\usepackage{xcolor}
\usepackage[numbers]{natbib}
\usepackage{float}
\usepackage{color,soul}
\usepackage{etoolbox}
\usepackage{ulem}
\usepackage{tikz}
\usepackage{url}
\usepackage{array}
\usetikzlibrary{positioning, fit, arrows.meta, shapes}
\usepackage[latin1]{inputenc}
\usepackage{booktabs}
\usepackage{caption}

\makeatletter
\patchcmd{\@makecaption}
  {\scshape}
  {}
  {}
  {}
\makeatletter
\patchcmd{\@makecaption}
  {\\}
  {.\ }
  {}
  {}
\makeatother
\def\tablename{Table}

\def\BibTeX{{\rm B\kern-.05em{\sc i\kern-.025em b}\kern-.08em
    T\kern-.1667em\lower.7ex\hbox{E}\kern-.125emX}}

\pgfdeclarelayer{background}
\pgfdeclarelayer{foreground}
\pgfsetlayers{background,main,foreground}

\begin{document}

\title{A comparison of machine learning surrogate models of street-scale flooding in Norfolk, Virginia}

\author{\IEEEauthorblockN{Diana McSpadden}
\IEEEauthorblockA{\textit{Data Science Department} \\
\textit{Thomas Jefferson National Accelerator Laboratory}\\
Newport News, VA 23606, USA \\
0000-0002-8520-1631}
\and
\IEEEauthorblockN{Steven Goldenberg}
\IEEEauthorblockA{\textit{Data Science Department} \\
\textit{Thomas Jefferson National Accelerator Laboratory}\\
Newport News, VA 23606, USA \\
0000-0002-5264-6298}
\and
\IEEEauthorblockN{Binata Roy}
\IEEEauthorblockA{\textit{Department of Civil and Environmental Engineering} \\
\textit{University of Virginia}\\
Charlottesville, VA, USA \\
0000-0002-2726-3340}
\and
\IEEEauthorblockN{Malachi Schram}
\IEEEauthorblockA{\textit{Data Science Department} \\
\textit{Thomas Jefferson National Accelerator Laboratory}\\
Newport News, VA 23606, USA \\
0000-0002-3475-2871}
\and
\IEEEauthorblockN{Jonathan L. Goodall}
\IEEEauthorblockA{\textit{Department of Civil and Environmental Engineering} \\
\textit{University of Virginia}\\
Charlottesville, VA, USA \\
0000-0002-1112-4522}
\and
\IEEEauthorblockN{Heather Richter}
\IEEEauthorblockA{\textit{School of Community and Environmental Health} \\
\textit{Old Dominion University}\\
Norfolk, VA, USA}
}
\maketitle

\begin{abstract}
 Low-lying coastal cities, exemplified by Norfolk, Virginia, face the challenge of street flooding caused by rainfall and tides, which strain transportation and sewer systems and can lead to property damage. 
 While high-fidelity, physics-based simulations provide accurate predictions of urban pluvial flooding, their computational complexity renders them unsuitable for real-time applications. 
 Using data from Norfolk rainfall events between 2016 and 2018, this study compares the performance of a previous surrogate model based on a random forest algorithm with two deep learning models: Long Short-Term Memory (LSTM) and Gated Recurrent Unit (GRU).
 This investigation underscores the importance of using a model architecture that supports the communication of prediction uncertainty and the effective integration of relevant, multi-modal features. 
\end{abstract}


\section{Introduction and Motivation}\label{sec:introduction}

Simulated street-scale predictions of water depth throughout rainfall events can provide insights to urban decision-makers, enabling them to understand and address potential disruptions in areas like transportation and emergency management.
High-resolution/high-fidelity, physics-based hydrology models are impractical for real-time decision support due to long computation times. 
Additionally, realistic simulation models such as Two-dimensional Unsteady Flow (TUFLOW) models \cite{wbm2016tuflow}, the source of simulated  \textquotedblleft truth\textquotedblright \ used here, require extensive calibration and parametrization for each location where they are used.

The ability to predict street-scale flooding in real-time offers significant practical benefits for city officials, emergency management agencies, transportation authorities, and community members. 
It is of such interest that in 2020 a pilot project was funded to provide flooded road condition information directly to drivers via the Waze for Cities program \cite{FloodMapp}. 
In addition to direct-to-driver solutions, for emergency planning, a fast simulation of street-scale flooding must be able to run at timescales that allow for \textquotedblleft what-if\textquotedblright \ analysis.
In the context of machine learning (ML) decision support applications, experts require a surrogate model that provides well-calibrated predictions with uncertainties to optimize different planning scenarios. 
These scenarios may require running the optimization multiple times to understand the impact of surrogate model parameters and inputs on water level predictions. 

This work applies Recurrent Neural Networks (RNNs), which are ML models capable of remembering previous sequences and forecasting lead time predictions. 
RNN surrogate models for an urban mesh 3D hydrograph of flooding predictions have yet to be explored and are a plausible model architecture to investigate.
This work builds on Zahura et al. \cite{RF_streetLevelFloodingModel}, which demonstrated a 3,000-time speedup compared to the TUFLOW physics-based model. 
Zahura et al. employed the random forest (RF) algorithm for predicting street flooding. 
In contrast, this study compares two RNN models: the Long short-term memory (LSTM) model \cite{hochreiter1997long} and the Gated Recurrent Unit (GRU) model \cite{GRU-cho2014learning}. 
ML surrogate models have been utilized for various applications related to urban flooding prediction. 
These include probabilistic flood occurrence prediction \cite{li2020hybrid}, forecasting flood risk indices or susceptibility \cite{MOTTA2021102154, DARABI2021126854, FANG2021125734}, estimating total accumulative water overflow during an event \cite{kim2020urban}, and predicting maximum water depth for flood events \cite{WU2020137077, BERKHAHN2019743, guo2021data, LOWE2021126898}.

Another noteworthy work by Zhang et al. \cite{ZHANG2023129499} introduces an LSTM model with an attention mechanism and weighted mean squared error loss (ALSTM-DW) to address street-scale flooding in three specific locations in Shenzhen, China. 
The study presents promising results, particularly considering the minute scale at which the model performs predictions. 
In contrast to this work, ALSTM-DW trains a separate model for each of the three street locations rather than considering the street segment mesh as a whole. This distinction is important as this work aims to produce a deep learning surrogate model capable of receiving multi-modal inputs and producing uncertainty quantification (UQ). Accepting diverse input modalities, such as images, LIDAR, and vector data, and developing well-calibrated models with uncertainties for flooding are future endeavors; however, this work emphasizes such capabilities are essential for a generalizable decision-support flooding model able to be used in diverse locations and climate scenarios.

Flooding events examined in both Zahura et al. and this work include those categorized as Nuisance Flooding (NF), in contrast to flooding caused by extreme events or disasters. 
The NF definition is standard but one of comparison rather than criteria. 
Moftakhari et al. \cite{whatisnuisanceflooding} propose a definition of NF based on transportation, public health, and property damage impacts with NF lower and upper thresholds for water velocity and depth (0 - 3 m/s and 1 - 3 cm, respectively). 
NF may be referred to as \textquotedblleft clear sky flooding\textquotedblright or \textquotedblleft sunny day flooding,\textquotedblright; however, NF encompassed all causes of low levels of inundation, including fluvial (river), pluvial (rain), and tidal.

In this work, Section \ref{sec:Norfolk} provides a comprehensive description of the study area, Norfolk, Virginia, including its vulnerability and relevance to coastal, pluvial urban flooding. 
Section \ref{sec:data} elaborates on the study data, encompassing both temporal and spatial features. Following this, Section \ref{sec:data_manageandprep} outlines the data management and preparation procedures. Subsequently, Section \ref{sec:methods} provides a review of the machine learning methods employed. Lastly, a presentation of results and a discussion of the importance  of a deep learning surrogate model architecture are presented.

\section{Coastal Urban Flooding in Norfolk, VA}\label{sec:Norfolk}
As the second most populous city in the Hampton Roads region with a total population of over 1.7 million, Norfolk, Virginia is home to a diverse population and is a major economic and cultural center. 
Additionally, its position as the site of the world's largest naval base, the deepest water harbor on the US East Coast, and a North Atlantic Treaty Organization headquarters makes it a critical location for national and international interests.
Norfolk has seen a significant increase in sea level rise (SLR) relative to land and increasing mean sea level (MSL)\cite{sweet2014sea}. 
Vertical land subsidence in the lower Chesapeake Bay combines with global SLR and results in the Norfolk community's increasing vulnerability to flooding during storm tides and rainfall events \cite{sweet2014sea, boon2010chesapeake}. 
Norfolk is the second most vulnerable community to coastal flooding in the US, behind New Orleans \cite{fears2012built}. 
This risk is striking when considering Norfolk's seven miles of Chesapeake Bay beachfront, additional 137 miles of shoreline along rivers and lakes, the world's largest naval base, and working waterfronts on every major waterway in the city, which include the Elizabeth River, the James River, the Chesapeake Bay and Little Creek \cite{virginia_waterfront_plan}.

Norfolk is particularly vulnerable to coastal flooding and has seen a 325\% increase in NF since 1960 \cite{burgos2018future}. 
Burgos et al. goes on to state that \textquotedblleft... this flooding will continue to increase in frequency with time, with a potential for well over 200 flood events in the year 2049 [in Norfolk].\textquotedblright \ 
Not only will the number of events increase, but their characteristics will also diverge from what is seen today. 
For example, Burgos et al. describe how at some point in the future, due to climate change and SLR, Norfolk will experience NF solely from high tide events, without wind or rain contributions. 
The availability of the Norfolk data set, the previous RF surrogate model, and the urgency of the street-level flooding issue in the study area motivate this work.

\section {Data}\label{sec:data}

Zahura et al. provided data for the study area and the rainfall event data, including street segment selection and feature engineering as described in the following subsections. Additionally, they offered pre-processing instructions to filter street segments pertaining to Norfolk underpasses where the TUFLOW model lacks sufficient information on pump stations for reliable physics-informed predictions. 
Consequently, these street segments were filtered from both Zahura et al.'s dataset and this work. 
Nevertheless, out of the total of 16,923 street segments, there are nine segments that remain excluded from Zahura et al.'s dataset, which comprises 16,914 street segments. Although the reason behind this discrepancy in street segment inclusion could not be determined during the analysis, the resulting inconsistency has a negligible impact on the outcomes given the substantial number of street segments involved.

\begin{figure}
    \centering
    \includegraphics[width=0.45\textwidth]{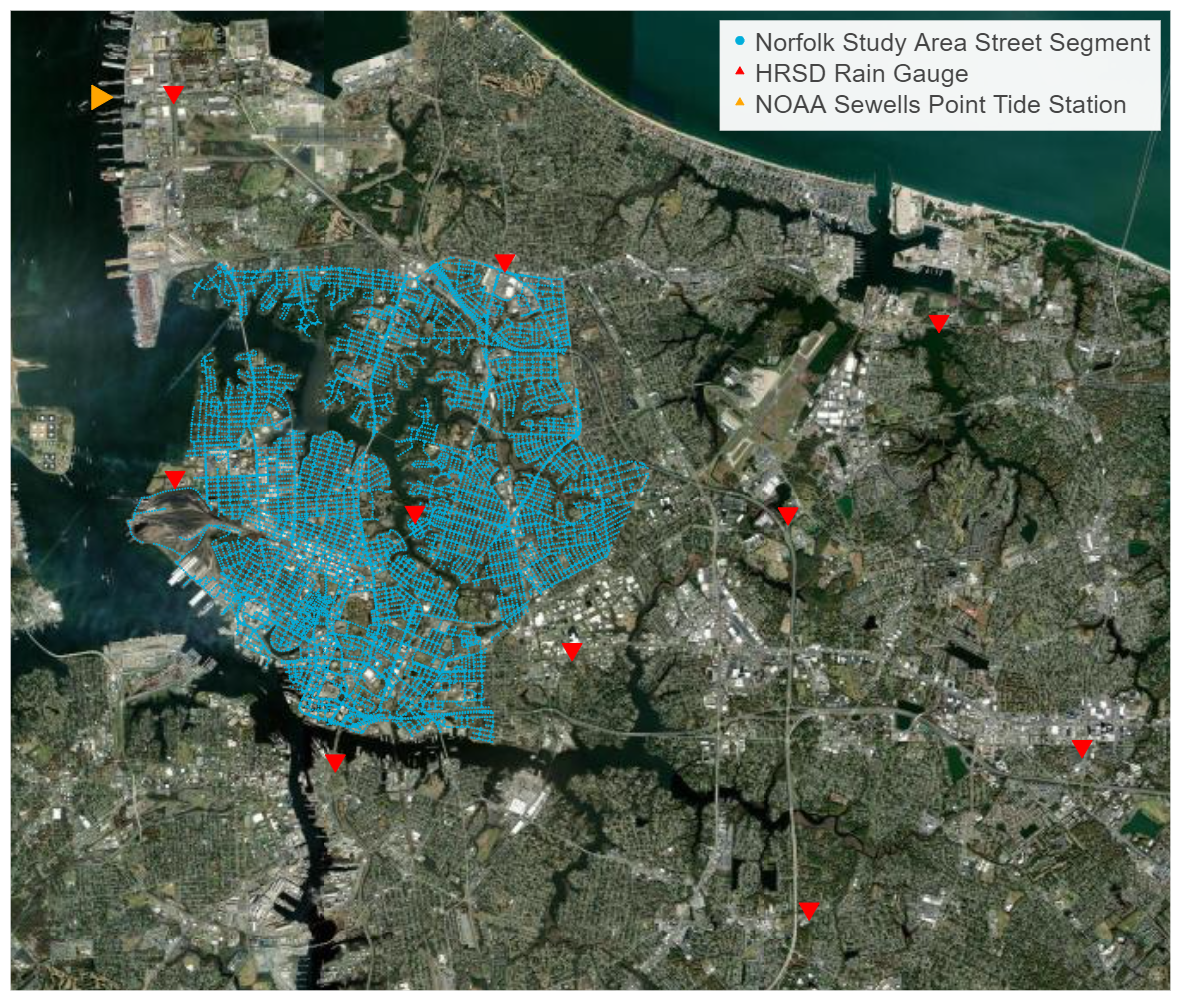}
    \caption{The locations of the Norfolk study area's 16,923 street segments, 10 Hampton Roads Sanitation District rain gauges, and the U.S. National Oceanic and Atmospheric Administration Sewell's Point Tide station gauge are shown on an area map.}
    \label{fig:NorfolkStudyArea}
\end{figure}

\begin{figure}
    \centering
    \includegraphics[width=0.45\textwidth]{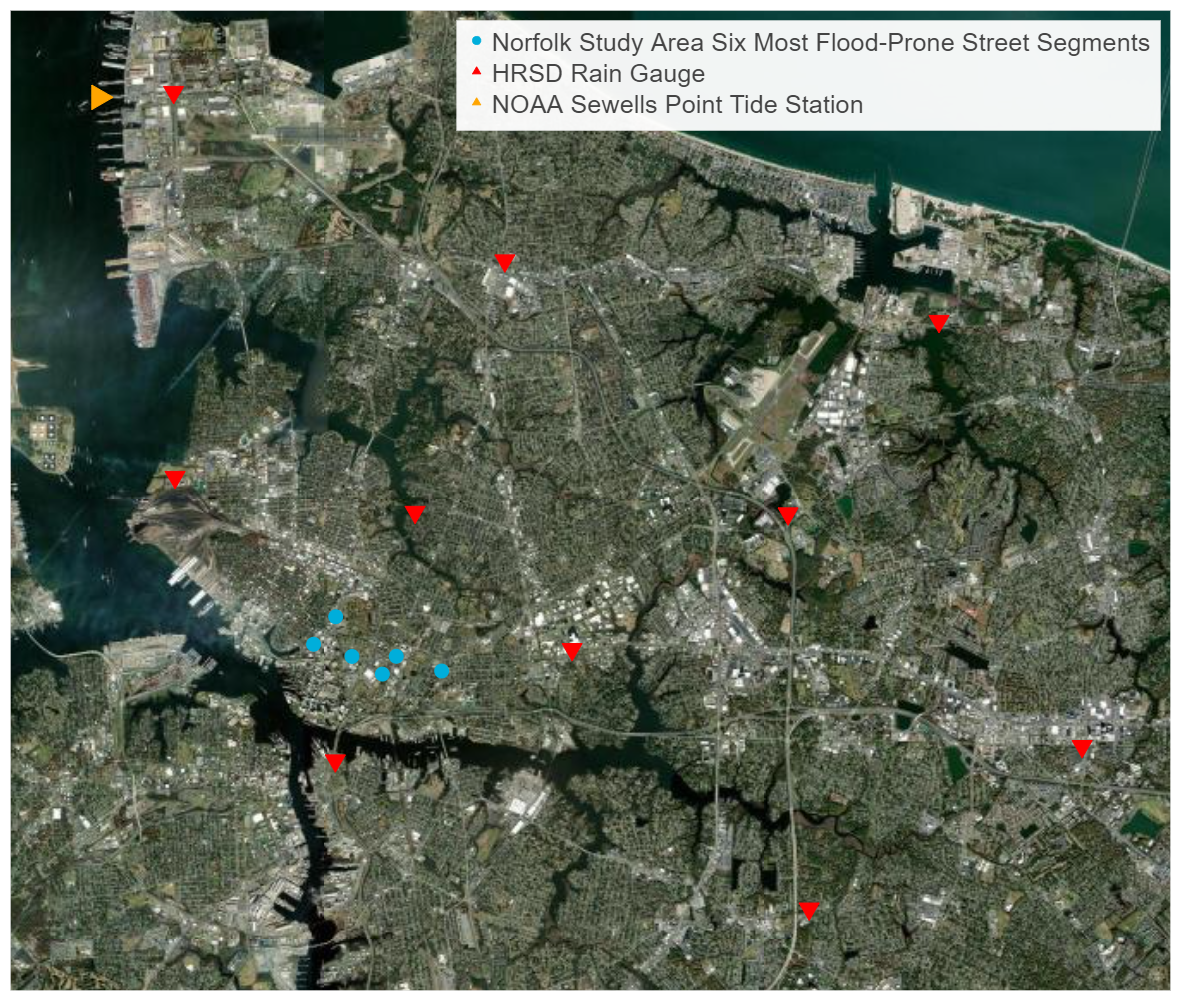}
    \caption{Mini study area of six most flood-prone street segments of the Norfolk, VA study area, all located on the south-central section of the study area.}
    \label{fig:SixFloodProne}
\end{figure}

\subsection{Study Areas}\label{sec:studyareas}
Fig. \ref{fig:NorfolkStudyArea} illustrates the entire Norfolk study area's 16,923 street segments covering nearly 700 km of roadways in the region. Each street segment represents a 50 m $\times$ 7.2 m road segment. 
The average road lane width in the United States is 3.6 m \cite{usdot}. This informed the selection of the 7.2 m road segment width. 

Norfolk's six most flood-prone locations during the period 2010-2018 were identified using the city's System to Track, Organize, Record, and Map (STORM) \cite{STORM}. 
These locations, shown in Fig. \ref{fig:SixFloodProne}, are a convenient sample for a neural architecture search and help to gauge model performance. Notably, these street segments are situated at lower elevations, with a mean elevation of 1.7 m, and a max elevation of 9.7m  compared to the full study area's mean elevation of 2.8 m and max elevation of 12.5 m. Zahura et al. labeled the RF surrogate model trained using the mini-study area as Model 1, while the RF surrogate model trained using the entire study area is referred to as Model 2.

\subsection{Spatial Data} \label{sec:geodata}
For each street segment, the spatial characteristics are described with three topographical features: elevation (ELV), topographical wetness index (TWI) \cite{beven1979physically}, and depth to water (DTW) index \cite{murphy2007mapping}. 
TWI is a measure of a location's accumulation of water runoff from surrounding areas. 
Under normal circumstances, lower elevations with lower slopes are more likely to retain water and have higher TWIs and higher elevations, while steeper slopes, will shed water and have lower TWIs. 
DTW is an estimate of the depth to the water table for the street segment. 
One-meter Digital Elevation Model (DEM) was obtained from the National Elevation Dataset, a product of the U.S. Geological Survey\cite{USGS_map}. 
Additionally, some supporting geographical information, unused by the ML models, was also provided: the latitude and longitude of the middle point for each street segment, the street name, and a street segment id.

\subsection{Temporal Data}\label{sec:raineventdata}

\begin{figure}
    \centering
    \includegraphics[width=0.4\textwidth]{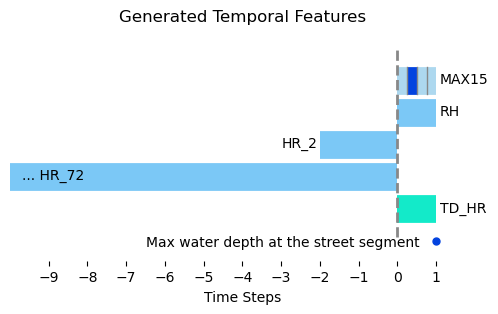}
    \caption{Features engineered from the Hampton Roads Sanitation District (HRSD) rain gauges and the TUFLOW high-fidelity simulation are shown. MAX15 is the cumulative total rainfall for the 15-minute interval with the greatest precipitation within the previous hour. Zahura et al. generated MAX15, RH, HR\_2, HR\_72 using a distance-weighted interpolation from the HRSD rain gauges. RH, HR\_2, HR\_72 are described in Table \ref{tab:input_features}. 'Max water depth at the street segment' is the maximum water depth for a street segment predicted by TUFLOW at each time step, in this case, hourly.}
    \label{fig:generated_temporal_features}
\end{figure}

Fifteen-minute rainfall measurements were provided from 10 Hampton Roads Sanitation District (HRSD) observation sites, as shown in Fig. \ref{fig:NorfolkStudyArea} and Fig. \ref{fig:SixFloodProne}. 
These measurements were then aggregated to create four statistical rainfall features, detailed in Table \ref{tab:input_features}. Zahura et al. employed inverse distance-weighted interpolation to account for the spatial variability of rainfall across the region and provide values for each street segment.
In Fig. \ref{fig:generated_temporal_features}, the engineered temporal rainfall features are presented alongside the TUFLOW predicted water depth, illustrating their temporal alignment.
Among the calculated rainfall features, the forecasting of maximum 15-minute rainfall within an hour posed a challenge as no suitable forecasting data source is available. 
Consequently, this analysis involves comparing the performance of each model architecture with and without the inclusion of the MAX15 feature. 
This approach serves a twofold purpose: firstly, it facilitates a performance comparison between RNN models and the RF model when utilizing the MAX15 feature, and secondly, it allows us to explore the effectiveness of an RNN model operating in a decision-support scenario when the MAX15 feature is unavailable.  
The hourly sample rate is the highest precision available from available forecast data sources, such as the National Weather Service API \cite{NWSAPI}, and allows for direct comparison of results.

\begin{table}
    \small
    \centering
    \begin{center}
        \caption{\textbf{Surrogate Model Inputs} \\
        RNN models are trained and evaluated with and without the MAX15 feature.}
    \end{center}
    \begin{tabular}{ p{2.85cm} p{1.2cm} p{.5cm} p{2.2cm} }
        \toprule 
        \textbf{Description} & \textbf{Abbrev.}  & \textbf{Unit} & \textbf{Variability} \\
        \midrule 
        Total hourly rainfall                   & RH       & mm & spatial, temporal  \\
        Max 15-minute rainfall in an hour       & MAX15*   & mm & spatial, temporal  \\
        Cumulative rainfall previous 2 hours    & HR\_2    & mm & spatial, temporal \\
        Cumulative rainfall previous 72 hours   & HR\_72   & mm & spatial, temporal  \\
        Hourly tide level                       & TD\_HR   & m  & temporal  \\
        Mean street segment elevation           & ELV      & m  & spatial  \\
        Topographic wetness index               & TWI      & --- & spatial \\
        Depth to water index                    & DTW      & cm & spatial \\
        \bottomrule 
    \end{tabular}
    \label{tab:input_features}
\end{table}

Hourly tide measurements were obtained for the simulated rainfall event dates and times. 
The hourly tide levels for the NOAA's Sewell's Point station are referenced to the North American Vertical Datum (NAVD 88) \cite{NAVD88}, the vertical control datum used for North American tidal measurement. 
The Sewell's Point tide monitoring station location is indicated in Fig. \ref{fig:NorfolkStudyArea} and Fig. \ref{fig:SixFloodProne}.

The maximum water depth for each street segment was predicted by the high-resolution 1-D/2-D physics-based simulation, TUFLOW, for each hour throughout all rainfall events. The TUFLOW surrogate model includes building footprints and heights for large commercial buildings greater than 500 m$^2$ and the Norfolk storm pipe infrastructure.
These water depth predictions are used as the target values.

Many events span multiple days, resulting in 20 total days of rainfall. 
Table \ref{tab:rainfallevents} shows the list of events, duration in hours, and total maximum hourly rainfall averaged over the study area's street segments.
Each event begins two to three hours/time steps before a storm and ends several hours/time steps after rainfall ceases. 
RNN models use a selected number of time steps as input when predicting the current time step. The short, five-hour duration of the August 20, 2018 event, shown in Table \ref{tab:rainfallevents}, results in the event being ill-suited for RNN training. Thus, this event is not used in this work.

\begin{table}
    \caption{\textbf{Rainfall Events}}
    \centering
    \begin{tabular}{@{}>{\raggedleft\arraybackslash}p{2.7cm} >{\centering\arraybackslash}p{1.cm} >{\centering\arraybackslash}p{2.4cm} >{\centering\arraybackslash}p{0.85cm}@{}}
        \toprule
        \textbf{Event Dates (mm/dd/yyyy)} & \textbf{Duration (hrs)} & \textbf{Total max hourly rainfall (street segments average) (mm)} & \textbf{Dataset} \\
        \midrule
        06/05/2016-06/06/2016 & 28 & 78.7 & Train \\
        07/30/2016-07/31/2016 & 34 & 200.7 & Train \\
        08/09/2016 & 16 & 37.9 & Train \\
        09/02/2016-09/03/2016 & 28 & 109.8 & Train \\
        09/19/2016-09/21/2016 & 60 & 329.9 & Train \\
        10/08/2016-10/09/2016 & 37 & 328.5 & Train \\
        01/01/2017-01/02/2017 & 23 & 55.2 & Train \\
        07/14/2017-07/15/2017 & 22 & 132.1 & Train \\
        08/07/2017-08/08/2017 & 34 & 117.9 & Train \\
        08/28/2017-08/29/2017 & 25 & 107.0 & Train \\
        10/29/2017-10/30/2017 & 29 & 57.3 & Test \\
        05/06/2018 & 24 & 82.8 & Test \\
        05/28/2018-05/29/2018 & 26 & 98.6 & Test \\
        06/21/2018-06/23/2018 & 37 & 96.3 & Train \\
        07/30/2018 & 11 & 70.0 & Train \\
        08/11/2018 & 24 & 58.2 & Test \\
        \sout{08/20/2018} & 5 & \sout{61.3} & \sout{Train} \\
        \bottomrule
    \end{tabular}
    \label{tab:rainfallevents}
\end{table}

\subsection{Training and Test Rainfall Events}\label{sec:traintest}
\begin{figure}
    \centering
    \includegraphics[width=0.33\textwidth]{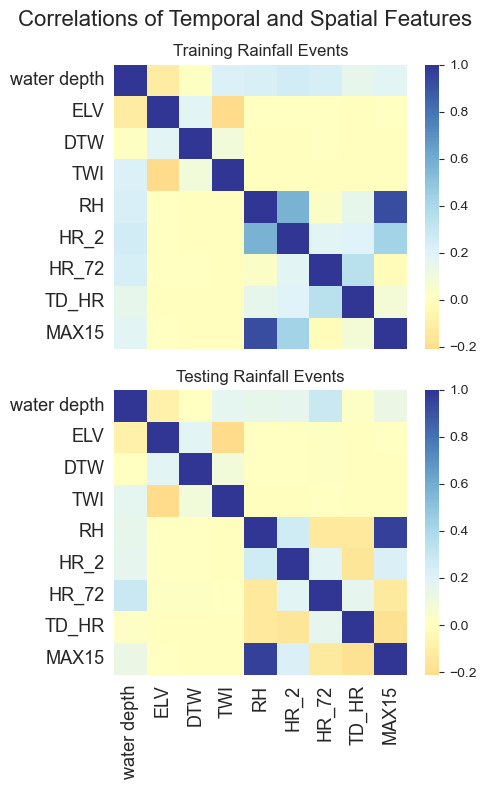}
    \caption{For the train and test rainfall events, correlation heatmaps of the target water depth, and the spatial and primary temporal features. Water depth is the target variable. Table \ref{tab:input_features} describes the input features for the surrogate model. The Pearson correlation coefficient was used to calculate the correlation.}
    \label{fig:temporal_spatial}
\end{figure}

Table \ref{tab:rainfallevents} displays the assignment of a rainfall event to either the training or test dataset according to Zahura et al.'s assignment.
Fig. \ref{fig:temporal_spatial} illustrates the correlations between the target variable, water depth, and the input features for the training and test events. 
Water depth exhibits only weak correlations with all input features; the heatmaps reveal that the problem cannot be addressed through simple regression using hourly rainfall alone. 
Also of note, there are variations in the correlations between the training and testing sets. 
Specifically, the correlation between RH and TD\_HR changes between the training and testing sets (Pearson's r: $0.15$ and $-0.14$, respectively). 
Furthermore, there is a disparity in the correlations of RH and HR\_2, which is moderately positively correlated (Pearson's r: $0.57$) in the training dataset but only weakly positively correlated (Pearson's r: $0.27$) in the testing dataset. 
Despite these differences in correlations, this work maintains the train/test event split to facilitate a direct evaluation in line with research objectives. 

\section{Data Management and Preparation}\label{sec:data_manageandprep}

\subsection{Relational data management}
The data from Zahura et al. study was shared as a collection of csv files, one for each of the rainfall events, each with columns for the environmental and geographical fields for each of the street segments. 
This resulted in the repetition of geographic features, which are static by event, and the repetition of tide measurements, which are static by street segment for each hour. 
We opted to create a relational file structure. Storing the static geographic data for each street segment in a street segment specific csv file, with a street segment ID. 
Storing the hourly tide levels, which are the same for every street segment, but change over time in a tidal specific csv file with a date-time ID. 
Finally, storing the rainfall and tide measurements, which vary both spatially and temporally, in a weather csv, with both street segment and date-time IDs. 
This relational data structure allowed for a compact representation of the data and for simple data querying and loading for selections of street segments and rainfall events.

\subsection{Data preparation}

Data are queried from the relational data structure for the studied street segments; the training, validation, and test events are stored in separate dataframes. 
Scikit-learn's \cite{scikit-learn} Standard Scaler is used to scale parameters using the mean and standard deviation of the training set for the training, validation, and testing data sets being evaluated. 
Variables include non-temporal/spatial and temporal predictors, which inspired using a model with an input branch of layered LSTM or GRU cells to learn temporal features. Temporal inputs include RH, HR\_2, HR\_72, TD\_HR, and optionally the MAX15 feature. 
Temporal variables are structured to match the required input to the LSTM or GRU layer: [samples, timesteps, features], ensuring that the boundaries between events were respected. Static variables, i.e. the ELV, TWI, and DTW for each street segment, are provided to the model in a separate input branch in the shape [samples, features].

\section{Machine Learning Methods}\label{sec:methods}

This work utilizes RNNs, a model architecture designed to learn from sequential data via backward connections, enabling learning from previous time steps. 
During the training, each neuron in the RNN receives predictors for the current time step and predictand(s) from a chosen number of preceding time steps. 
This is in contrast to basic feed-forward artificial neural networks. 
The key advantage of RNNs lies in their ability to capture temporal order and learn temporal dependencies within sequential data \cite{HEWAMALAGE2021388}. However, it is important to note that basic RNNs are known to suffer from unstable gradient problems \cite{geron2022hands} due to the same weights being used at each time step. Additionally, some information is lost at each time step, which is particularly problematic for long sequences where the initial input may be forgotten. To address these issues, LSTM and GRU cells have been introduced. In this work, the Keras \cite{chollet2015keras} implementations of these cells are implemented.

\subsection{LSTM and GRU cells}
LSTM cells were introduced by Hochreiter and Schmidhuber in 1997 \cite{hochreiter1997long,geron2022hands}, and have been successful in modeling and predicting data with temporal dependencies. 
During the training process, input for each time step is processed through multiple gates that regulate the retention of information for long-term and short-term patterns. 
The current input and the previous short-term state are input to four fully connected dense layers, determining what is stored in long-term memory, what is forgotten, what is produced as the short-term state, and what is output as the next time step. 
The previous long-term state navigates a forget gate and selectively discards previous data that does not contribute to learning. LSTM cells mitigate the issue of vanishing gradients in RNNs through the use of these multiple gate controllers. LSTM cells improve traditional RNNs by preserving critical patterns within the long-term state for as long as they are useful.

GRU cells introduced by Cho et al. in 2014 \cite{GRU-cho2014learning}, generally perform equivalently to LSTM cells while simplifying the LSTM architecture by combining the input and forget gates into a single gate controller. 
When the gate controller outputs the value of $1$, the forget gate is open, and the input gate is closed. 
Conversely, when the gate controller outputs the value of $0$ the forget gate is closed, and the input gate is open.
Additionally, GRU cells have no explicit output gate. 
Instead, a gate controller determines which part of the previous time step is exposed to the main layer of the GRU cell \cite{geron2022hands}. 

\subsection{Architecture selection}
For the temporal aspect, one-hour and four-hour look-backs are considered. The four-hour look-back results in a lower mean absolute error (MAE) and root-mean-square error (RMSE) for both the LSTM and GRU models using the mini-study area as shown in Table \ref{tab:sixnode}. 
Ultimately, the model produces a single time step, i.e., one-hour look-ahead inference. The choice of one-hour time-step flooding predictions is deemed reasonable as all input features, except MAX15, are available via forecast.
The initial input branch of the model captures information from the dynamic spatial and temporal tidal and pluvial events by employing single or stacked LSTM or GRU layers. 
The activation function for the RNN layer was set as the hyperbolic tangent function (\textit{tanh}). 
L1 and L2 regularization are employed on the bias and kernel weights of these layers. 
The regularization factors for both L1 and L2 were set to $0.01$ to control the impact of the weights on the model's overall performance. 
The complete sequence of the RNN is then passed on to the subsequent layer.

A second input branch incorporates dense layers that extract features from the spatial inputs, namely ELV, TWI, and DTW. 
Subsequently, the outputs from the two branches are concatenated, and additional dense layers are utilized to extract relevant features from the combined temporal and spatial information. 
The $\tt{Nadam}$ \cite{dozat2016incorporating} optimizer function was used for training, and MAE was used for the loss function.

\begin{figure}
    \centering
    \includegraphics[width=0.3\textwidth]{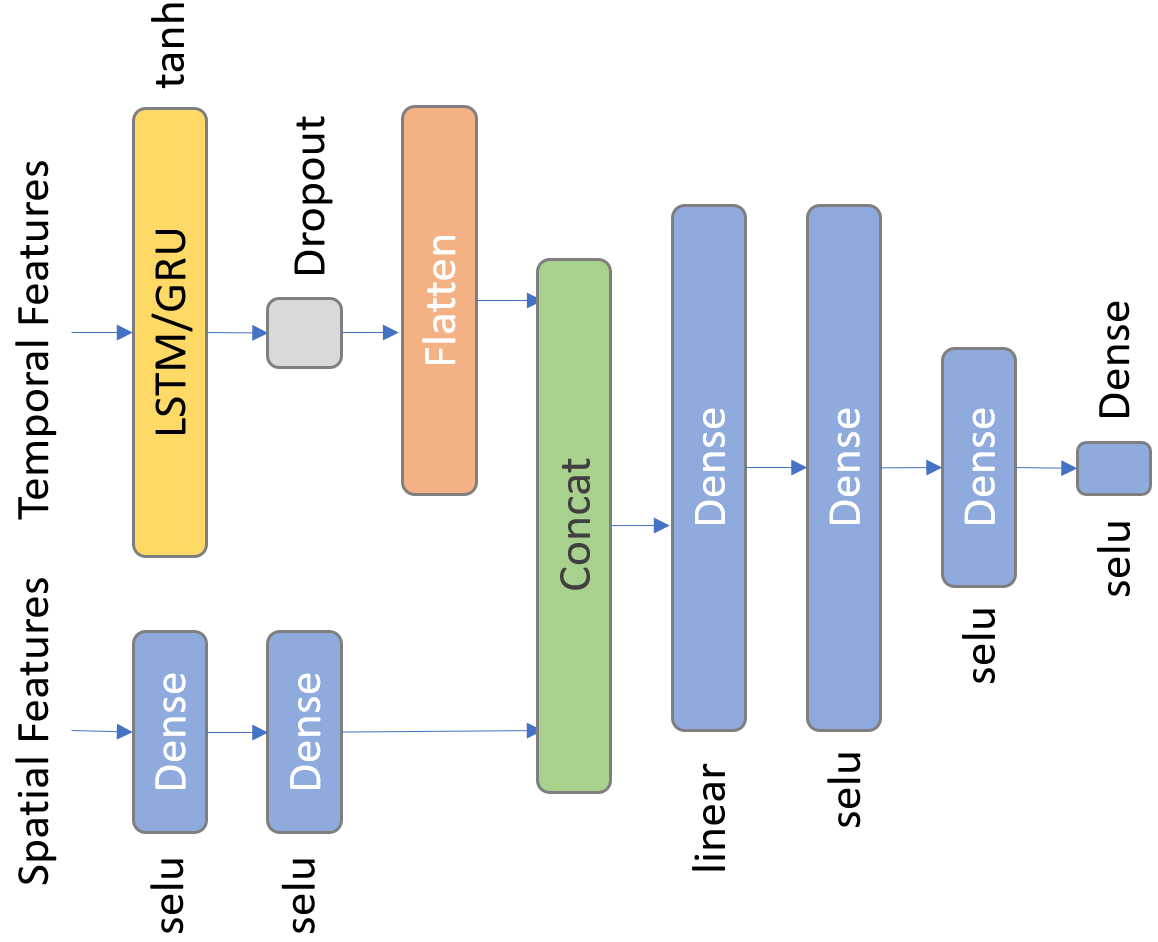}
    \caption{Diagram of the champion model architecture after the neural architecture search on the six flood prone street segments.}
    \label{fig:modeldiagram}
\end{figure}

\subsection{Neural architecture search}

To optimize the model's parameters, a grid search was conducted using the flood-prone street segment mini-study area defined in Section \ref{sec:studyareas}. 
The parameters evaluated are shown in Table \ref{tab:hpo_grid}. 
Throughout the process, model performance was tracked using MLFlow \cite{databricks2020mlflow}. 
For each model architecture evaluated, twelve models were trained by setting aside each of the training rainfall events as a validation event and using the remaining eleven events for training. The average MAE and RMSE of the four test rainfall events were then averaged for the twelve trained models. 
Table \ref{tab:sixnode} displays the results of the champion model architecture with the lowest MAE.

\begin{figure*}
    \centering
    \includegraphics[width=1.0\textwidth]{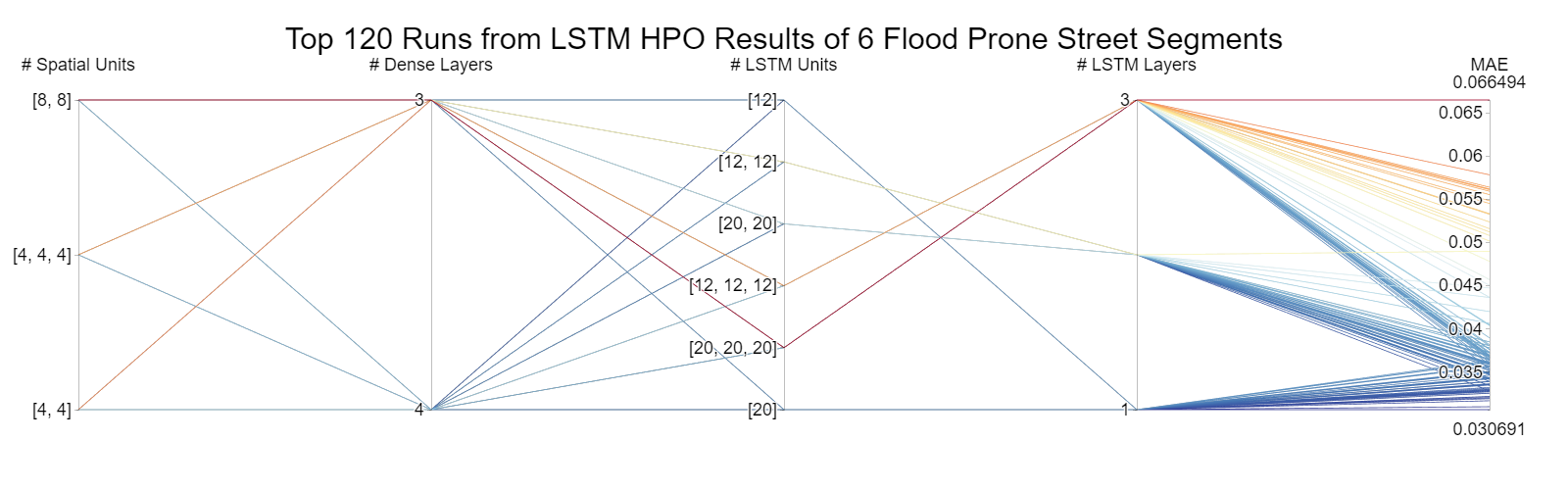}
    \caption{The parallel coordinate plot of the top 120 runs from the NAS of the mini-study area using a model using LSTM layers for temporal feature extraction.}
    \label{fig:hpo_pc}
\end{figure*}

\begin{figure}
    \centering
    \includegraphics[width=0.43\textwidth]{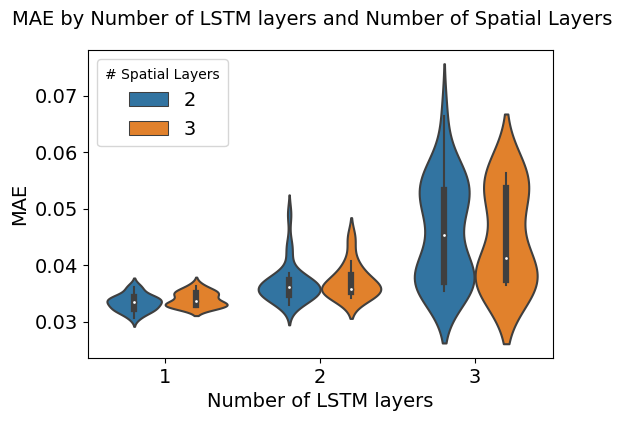}
    \caption{Distributions of the MAE for the top 120 runs from the NAS of the mini-study area using an LSTM-based RNN model. MAE distributions are shown for the number of LSTM layers included in the model and for the number of stacked dense layers used to extract features from the spatial inputs for a street segment: elevation, topographical wetness index, and depth to water.}
    \label{fig:hpo_MAE_spatial}
\end{figure}

\begin{figure}
    \centering
    \includegraphics[width=0.43\textwidth]{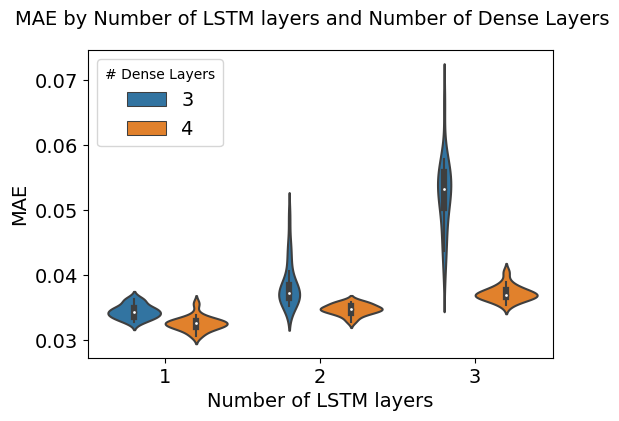}
    \caption{ Distributions of the MAE for the top 120 runs from the NAS of the mini-study area using an LSTM-based RNN model. MAE distributions are shown for the number of LSTM layers included in the model and for the number of stacked dense layers used to extract features from the concatenated outputs from the temporal and spatial model branches.}
    \label{fig:hpo_MAE_output}
\end{figure}

The parallel coordinate plot of the top-performing 120 runs from the Neural Architecture Search (NAS) is displayed in Fig. \ref{fig:hpo_pc}. 
Surprisingly, models with only a single LSTM layer performed better than models with 2 or 3 stacked LSTM layers. 
From Fig. \ref{fig:hpo_pc}, it is also clear that models with 4 dense layers after concatenating the temporal and spatial branches performed better than models with 3 dense layers. 
Fig.'s \ref{fig:hpo_MAE_spatial} and \ref{fig:hpo_MAE_output} display the distributions of MAE for different numbers of stacked LSTM layers and different numbers of dense layers in the spatial branch (Fig. \ref{fig:hpo_MAE_spatial}), or the layers after concatenation (Fig. \ref{fig:hpo_MAE_output}). 
These figures illustrate that even with the complete output sequence of the LSTM passed to the next layer, the stacking of LSTM layers did not result in lower MAE with the architectures attempted. 
Fig.'s \ref{fig:hpo_MAE_spatial} and \ref{fig:hpo_MAE_output} also illustrate that the variability of MAE was positively correlated with an increase in the number of stacked LSTM layers as the other model parameters were evaluated.

 \begin{table}
    \caption{\textbf{Neural architecture search parameters}}
    \centering
    \begin{tabular}{@{}>{\raggedleft\arraybackslash}p{3.5cm} >{\raggedright\arraybackslash}p{4.1cm}@{}}
        \toprule
        \textbf{Parameter} & \textbf{Values} \\
        \midrule
        \# of LSTM layers & 1, 2, or 3  \\
        Units per LSTM layer & 12 or 20  \\
        \# of spatial dense layers & 2 or 3  \\
        Units per spatial dense layer  & 4 or 8  \\
        Spatial dense layer activation & \textit{relu}, \textit{selu}, and \textit{linear}\\
        \# of dense layers & 3 or 4  \\
        Units per dense layer  & [64, 64, 1], [32, 32, 1], [32, 16, 1], [64, 64, 16, 1], [64, 32, 32, 1]  \\
        Dense layer activation & \textit{relu}, \textit{selu}, and \textit{linear}\\
        \bottomrule
    \end{tabular}
    \label{tab:hpo_grid}
\end{table}

To evaluate the full study area the best-performing architecture was selected from the NAS results on the mini-study area. 
The selected model architecture's input branch for temporal features includes one LSTM, or GRU layer with 20 units and uses the \textit{tanh} activation function. 
Again, a 4-time-step look back is used. 
The spatial feature branch includes two dense layers, each with four units, and each using the \textit{selu} activation function. Finally, the two branches concatenate into four dense layers with 64, 64, 16, and 1 units, and \textit{linear}, \textit{selu}, \textit{selu}, and \textit{selu} activation functions. Fig. \ref{fig:modeldiagram} illustrates the selected model architecture. 

\section{Results and Discussion}\label{sec:results}

As depicted in Table \ref{tab:full_results}, the findings align with the previous results.
Sections \ref{sec:Results_UQ} and \ref{sec:Results_MultiModal} reveal that a deep learning approach provides State-Of-The-Art (SOTA) ML surrogate model accuracy and propels research forward, steering closer to the future objective: an ML model that 1) is uncertainty aware, and 2) accepts multi-modal input that can include additional sensor data for model refinement, calibration, and generalization. 
First, a comparison is made between this work and Zahura et al.'s RF surrogate model.

\subsection{Results}\label{sec:Results_RFCompare}
In contrast to the RF method, RNN methods incorporate a \textquotedblleft look back\textquotedblright \ approach, where the model leverages a certain number of preceding time steps as input while predicting the output at the current time step. 
For this study, a four-time-step look-back period was selected.
Consequently, the initial four time steps of each rainfall event lack the necessary input data for prediction, resulting in the exclusion of these time steps for predictions. 
This introduces a challenge in comparing the results of RNN models and previous SOTA RF results in \cite{RF_streetLevelFloodingModel}, as the initial hours of each event typically exhibit lower rainfall, minimal flooding, and, generally, lower error. 
Furthermore, \cite{RF_streetLevelFloodingModel} investigated the use of sample weights, oversampling, and a combined approach in training their RF model. 
Zahura et al. employed a combined approach to report results on the entire study area. 
This analysis, however, chose not to utilize combined approaches involving sample weights and oversampling.

Tables \ref{tab:sixnode} and \ref{tab:full_results} compare these results to Zahura et al. for the mini-study area and the full study area, respectively. 
As shown in both tables, the results for all models are equivalent, and the average MAE for the four test events is less than 5 centimeters from the TUFLOW predicted water depth. 
The comparison of the full study area MAE vs. RMSE is noteworthy. MAE was the loss function used for model training for both Zahura et al. and this work. 
Table \ref{tab:full_results} demonstrates that RMSE is sensitive to the presence of the street segments with the highest flooding \cite{hyndman2006another}. 

\subsection{Discussion}\label{sec:discussion}

\subsubsection{Uncertainty quantification} \label{sec:Results_UQ}
Unlike the previous RF model, this work utilizes RNN approaches, providing the advantage of integrating techniques that support the inclusion of UQ alongside their predictions.
This addresses the need highlighted by Boomers and Hulscher in their study on "Neural networks for fast fluvial flood predictions: Too good to be true?" \cite{bomers2023neural}, where they emphasize the importance of forecast uncertainty in predictions from artificial neural network models for fluvial flooding. 
In this pluvial flooding scenario, uncertainty estimations are essential for providing reliable real-time support as they inform decision-makers of changing conditions that could cause the model to poorly predict outcomes.
In future investigations, two types of out-of-distribution (OOD) data hold particular interest for urban pluvial flooding: temporal OOD changes and spatial OOD changes.
As an example of temporal OOD data, climate change may alter sea level (tide) and the frequency and severity of rainfall events. 
The frequency of rainfall can potentially change the initial conditions for a rainfall event, e.g., HR\_72. 
The severity of the rainfall event would alter the distribution of all rainfall variables leading to model predictions that should be accompanied by related and well-calibrated uncertainty estimates.
Additionally, decision-makers may want to test the model on spatially OOD data such as new flood-prone locations (e.g. New Orleans).

\begin{table}
    \caption{\textbf{Results for six flood-prone street segments}\\
    The RF Model 1 from Zahura et al. is compared to this work, with and without the MAX15 feature. Results are measured in average meters of water for six street segments. MAE and RMSE are reported averaged over the four test events, and six street segments.}
    \centering
    \begin{tabular}{@{}>{\raggedleft\arraybackslash}p{1.4cm} >{\centering\arraybackslash}p{1.3cm} >{\centering\arraybackslash}p{1.25cm} >{\centering\arraybackslash}p{1.25cm} >{\centering\arraybackslash}p{1.25cm}@{}}
        \toprule 
        & \textbf{MAX15} & \textbf{look-back}  & \textbf{MAE} & \textbf{RMSE} \\
        \midrule
        RF Model 1 & Y & N/A & 0.036 m & \textbf{0.057 m} \\
        LSTM & Y & 1 &0.035 m & 0.085 m \\
        LSTM & N & 1 & 0.035 m & 0.085 m \\
        LSTM & Y & 4 &0.033 m & 0.071 m \\
        LSTM & N & 4 & 0.033 m & 0.069 m \\
        GRU  & Y & 1 & 0.034 m & 0.085 m \\
        GRU  & N & 1 &  0.034 m & 0.085 m\\
        GRU  & Y & 4 & 0.033 m & 0.072 m \\
        GRU  & N & 4 &  \textbf{0.031} m & 0.069 m\\
        \bottomrule 
    \end{tabular}
    \label{tab:sixnode}
\end{table}

\begin{table}
    \caption{\textbf{Results for 16,923 street segments}\\
    The RF Model 2 from Zahura et al. is compared to this work with and without the MAX15 feature. MAE and RMSE are averaged over the four test events and all street segments.}
    \centering
    \begin{tabular}{@{}>{\raggedleft\arraybackslash}p{1.4cm} >{\centering\arraybackslash}p{1.4cm} >{\centering\arraybackslash}p{0.9cm} >{\centering\arraybackslash}p{1.1cm} >{\centering\arraybackslash}p{1.1cm}@{}}
        \toprule 
        & \textbf{sample weight} & \textbf{MAX15}  & \textbf{MAE} & \textbf{RMSE} \\
        \midrule 
        RF Model 2  & Y &  Y & \textbf{0.047 m} & \textbf{0.084 m} \\
        LSTM & N & Y & 0.048 m & 0.085 m \\
        LSTM & N & N & \textbf{0.047 m} & 0.085 m  \\
        GRU & N & Y & \textbf{0.047 m} & \textbf{0.084 m}  \\
        GRU & N & N & \textbf{0.047 m} & \textbf{0.084 m}  \\
        \bottomrule
    \end{tabular}
    \label{tab:full_results}
\end{table}

\subsubsection{Enabling Multi-Modal Input}\label{sec:Results_MultiModal}
TUFLOW, the high-fidelity hydrology model, not only captures the complex interactions of one and two-dimensional free-surface flows from complex tidal and pluvial events but also includes an engine to model a wide range of structures, including large underground networks of pipe systems, spillways, pits, manhole inlets, estuaries, and user-defined structures \cite{wbm2016tuflow}. 
Future work may explore capturing the local natural and built environment within the representation of a street segment. 
A potential approach directly utilizes image representations of DEM inputs to the spatial branch of the model. 
However, it is worth exploring the possibility of achieving comparable performance through a reduced input data representation. 
Employing a condensed latent space representation of the stormwater pipe infrastructure and building footprint for each street segment holds promise as a future avenue of investigation. 
These research directions raise the question of training a deep learning model for continuous spatial flooding prediction in the urban context. 

\subsubsection{Enabling Neighboring Street Segments}\label{sec:Results_GNN}

An alternative approach to understanding urban street-scale flooding leverages the adjacency of neighboring street segments. 
In this work, TWI is the only feature provided to the model which captures characteristics of the surrounding area, including the slope. 
However, a Graph Neural Network (GNN) \cite{GNN, Yu_2018} can establish a graph structure representing the study area, with street segments as nodes, and edges indicating any relationship between them. 
The GNN would be trained to predict water depth at the individual street segment level.
A GNN architecture could capture the flow of water over the two-dimensional landscape throughout a flooding event.
Including these features may not significantly improve the results presented in this paper; however, they could capture key interactions that may make the solution generalizable to other areas.  
Determining the optimal GNN aggregation approach remains an open question.

\subsection{Summary}
This study's findings underscore the potential of deep learning models, which exhibit comparable performance to RF models while providing the additional benefits of UQ and the ability to incorporate multi-modal inputs. Our work demonstrates that ML surrogate models can accurately predict urban street-scale NF, with predictions deviating by only centimeters from those of high-fidelity, physics-informed hydrology surrogate models. 
The evaluation of a deep learning method that facilitates the incorporation of UQ techniques such as Gaussian process \cite{wilson2016DKL}, deep quantile regression \cite{koenker2005QRBook}, and Monte Carlo dropout \cite{gal2016dropout} is a first step towards addressing concerns expressed by hydrology and emergency planning experts of using ML surrogate models for flooding. 
Using an input branch specific to extracting features from spatial inputs proved important to the model's performance and facilitates research into multi-modal input. 
Moreover, a model architecture that allows for exploring climate and built-environment scenarios, combined with UQ, can be a powerful tool for informing emergency management and urban planning. This work aims to inspire and facilitate future research in this direction.

\section{Declaration of Generative AI and AI-assisted technologies in the writing process}
During the preparation of this work, the authors used ChatGPT-3.5 for copy editing. After using this service, the authors reviewed and edited the content as needed and take full responsibility for the content of the publication.

\section{Acknowledgments}
Jefferson Science Associates, LLC operated Thomas Jefferson National Accelerator Facility for the United States Department of Energy under U.S. DOE Contract No. DE-AC05-06OR23177.

This work was supported by the US DOE as LAB 20-2261.

Additional funding for this work was provided by the Hampton Roads Biomedical Research Consortium as part of the efforts associated with the Old Dominion University-Thomas Jefferson National Accelerator Facility Joint Institute for Advanced Computing on Environmental Studies.

\typeout{}
\bibliographystyle{unsrt}
\bibliography{main}

\end{document}